\documentclass[letterpaper, 10 pt, conference]{ieeeconf}  

\IEEEoverridecommandlockouts                              

\usepackage{amsmath} 
\usepackage{amssymb}  
\usepackage{multirow} 
\usepackage{threeparttable}  
\usepackage{booktabs} 
\usepackage{graphicx}
\usepackage[ruled,lined,linesnumbered]{algorithm2e}
\usepackage{algpseudocode}
\usepackage{newtxtext}
\title{\LARGE \bf
Pheromone-Focused Ant Colony Optimization algorithm for path planning
}

\author{Yi Liu$^{1}$  Hongda Zhang$^{1}$ Zhongxue Gan$^{1, \ast}$ Yuning Chen$^{1}$ Ziqing Zhou$^{1}$ 
  Chunlei Meng$^{1}$   Chun Ouyang$^{1, \ast} $
\thanks{$^{1}$College of Intelligent Robotics and Advanced Manufacturing, Fudan University, Shanghai, China}%
\thanks{$^{*}$Corresponding author}%
}

\begin{document}

\maketitle
\thispagestyle{empty}
\pagestyle{empty}

\begin{abstract}
Ant Colony Optimization (ACO) is a prominent swarm intelligence algorithm extensively applied to path planning. However, traditional ACO methods often exhibit shortcomings, such as blind search behavior and slow convergence within complex environments. 
To address these challenges, this paper proposes the Pheromone-Focused Ant Colony Optimization (PFACO) algorithm, which introduces three key strategies to enhance the problem-solving ability of the ant colony. First, the initial pheromone distribution is concentrated in more promising regions based on the Euclidean distances of nodes to the start and end points, balancing the trade-off between exploration and exploitation. Second, promising solutions are reinforced during colony iterations to intensify pheromone deposition along high-quality paths, accelerating convergence while maintaining solution diversity. Third, a forward-looking mechanism is implemented to penalize redundant path turns, promoting smoother and more efficient solutions. These strategies collectively produce the focused pheromones to guide the ant colony's search, which enhances the global optimization capabilities of the PFACO algorithm, significantly improving convergence speed and solution quality across diverse optimization problems. The experimental results demonstrate that PFACO consistently outperforms comparative ACO algorithms in terms of convergence speed and solution quality.

\end{abstract}

 \section{INTRODUCTION}
\label{sec:intro}  
Ant colony optimization (ACO) algorithms~\cite{dorigo2019ant}, inspired by the cooperative foraging behavior of ants, are praised for their distributed intelligence, parallelism, positive feedback, and robustness.
With the widespread application of ACO algorithms~\cite{liu2023path}, researchers gradually found that ACO algorithms suffer from drawbacks such as low search efficiency, slow convergence, stagnation, and the tendency to converge to local optima. 
As a result,
several improvements to ACO have been proposed~\cite{cao2023random}. For example, variants such as elite ant colony optimization (EliteACO)~\cite{bullnheimer1999new}, which accelerates convergence by modifying pheromone updates based on the best solution found, and max-min ant colony optimization (MMACO)~\cite{stutzle2000max}, designed to mitigate premature stagnation by restricting pheromone concentration, have been widely adopted. 

In recent years, ACO algorithms have been widely used for path planning~\cite{zhang2024joint, guan2025real}, whose objective is to obtain an optimal and collision-free path from the origin to the destination.
For instance, GSACO~\cite{li2022ant}, based on an adaptive greedy strategy, dynamically adjusts the preference levels of ants during the search process. By leveraging this greedy strategy, GSACO encourages ants to explore regions with higher pheromone concentrations. However, experimental results indicate that this approach makes the colony prone to local optima.
Other researchers~\cite{skinderowicz2022improving} tried to control the degree of variation between newly constructed solutions and previously selected solutions to implement a more focused search, which ultimately preserves the quality of existing solutions while identifying potential improvements. Nevertheless, directly restricting the search scope in this way increases the risk of premature convergence to local optima.
Additionally, IHMACO~\cite{liu2023improved} incorporates four major strategies to enhance algorithm performance, featuring 11 adjustable parameters. However, its application to different problem instances necessitates a corresponding parameter configuration. Moreover, the execution time of IHMACO is long, limiting its practicality in real-world applications. This parameter tuning process requires experienced researchers to analyze the specific characteristics of each instance and make appropriate adjustments. Due to the uniqueness of different problem instances, such ACO algorithms struggle to generalize effectively in complex environments.
At the same time, the increasing complexity of improvements introduced by researchers in this field poses challenges for other scientists attempting to replicate and implement these strategies~\cite{10820101}. This complexity hinders the broader adoption of these algorithms. For example, PF3SACO~\cite{zhou2022parameter} employs a dynamic parameter adjustment mechanism integrating particle swarm optimization (PSO)~\cite{zhang2024joint} and a fuzzy system~\cite{chen2017robust}, which accelerates convergence and enhances accuracy and stability. However, its high computational complexity limits its practical feasibility.

To improve the solution quality and the speed of convergence of ant colony in path planning, this paper proposes Pheromone-Focused Ant Colony Optimization (PFACO) algorithm. To enhance search efficiency, an initial pheromone matrix is constructed based on environmental information to focus pheromone distribution on promising solutions, and achieving an effective balance between exploration and exploitation. 
From the perspective of population-level optimization, the algorithm adopts a promising solution reinforcement strategy inspired by biological evolution. Specifically, individuals generating high-quality solutions are propagated to subsequent generations, increasing their influence within the colony, while individuals producing inferior solutions have a higher probability of elimination. This reinforcement mechanism accelerates convergence through intensified pheromone reinforcement on promising paths while simultaneously preserving diversity.
Furthermore, at the individual level, each ant is endowed with short-term memory capabilities to enable localized solution refinement. This strategy reduces unnecessary path deviations by imposing penalties on excessive turning points, thus ensuring smoother and more efficient trajectories. 
Collectively, the proposed strategies allow the colony's pheromone to concentrate in regions close to optimal or near-optimal solutions, significantly enhancing the overall optimization performance of the algorithm.

The remainder of this paper is organized as follows: Section~\ref{sec:PRELIMINARIES} provides an overview of the traditional ant colony algorithm. Section~\ref{sec:PFACO} introduces the details of PFACO. The experiments are conducted, and the results are analyzed in Section~\ref{sec:experiments}. Finally, Section~\ref{conclusion} presents the conclusions.

\section{PRELIMINARY}
\label{sec:PRELIMINARIES}

The ACO algorithms are inspired by the cooperative behavior of ant colonies, which employs a dynamic pheromone updating mechanism and utilizes the number of ants $M$ to iteratively search $K$ times and approach the optimal or near-optimal solution.

This section presents the basic process of ACO~\cite{dorigo1996ant}.
In the initial iteration, each ant randomly selects an action from the available set, with all options having equal probability. After all ants complete their searches within an iteration, the pheromone concentrations are updated accordingly. The pheromone update is governed by the following~(\ref{eq:acoupdatepheromone}):
\begin{equation}\label{eq:acoupdatepheromone}
    \tau_{ij} = (1-\rho) \cdot \tau_{ij} + \sum_{m=1}^{M} \Delta \tau_{ij}^{m} , \  0<\rho<1
\end{equation}
where $M$ represents the maximum number of ants, $\rho$ denotes the pheromone volatility factor, $\tau_{ij}$ signifies the pheromone level between nodes $i$ and $j$, and $\Delta \tau_{ij}^{m}$ corresponds to the pheromone deposited by ant $m$ on the route from $i$ to $j$.
\begin{equation}\label{eq:deltatau}
\Delta \tau_{ij}^{m} = \begin{cases}
 Q/L_{m},  & \text{condition}\  {\mathrm{a}} \\
 0, & otherwise \\
\end{cases}
\end{equation}
$^{\mathrm{a}}$If ant $m$ visited edge (i,j) in its tour.

where $Q$ is a hyperparameter, and $L_{m}$ is the path length conducted by ant $m$.

In subsequent iterations, the action of the ant is guided by a state transition rule influenced by the pheromones deposited in previous iterations. This rule is defined as follows:
\begin{equation}\label{eq:acotransfer}
p_{ij}^m  = \begin{cases}
 \frac{ \tau_{ij}^{\alpha } \cdot \eta_{ij}^{\beta} } { \sum_{l\in allowed_{m} }^{} \tau_{il}^{\alpha } \cdot \eta_{il}^{\beta}} ,& if \ j \in allowed_{m} \\
 0, & otherwise \\
\end{cases}
\end{equation}

where $\alpha$ and $\beta$ are hyperparameters that regulate the balance between pheromone concentration and the heuristic function. 
The heuristic function $\eta$ is defined as the reciprocal of the distance between nodes $i$ and $j$ ($d_{ij}$), given by:
\begin{equation}\label{eq:acoheristic}
    \eta_{ij}=1/d_{ij}
\end{equation}
$allowed_{m}$ is defined as the set of neighboring nodes that ant $m$ can visit from its current position at node $i$.

\section{Pheromone-Focused Ant Colony Optimization algorithm (PFACO)}
\label{sec:PFACO}
This section introduces the details of PFACO algorithm. 
PFACO enhances the concentration of pheromones around optimal or near-optimal solutions through three strategies: Adaptive Distance Pheromone Initialization (ADPI), Promising Solutions Pheromone Reinforcement Strategy (PSPRS), and Lookahead Turning Optimization Strategy (LTOS). The pseudo-code of PFACO as shown in Algorithm~\ref{alg:PFACO}.

\subsection{Adaptive Distance Pheromone Initialization (ADPI) }
In traditional ACO algorithms, the initial pheromone concentrations at all positions in the environment are either identical (typically set to integer values such as 0 or 1) or determined based on the distances between nodes (like Equation (\ref{eq:acoheristic})). This uniform initialization results in high randomness in the ants' action selection during the early exploration phase, thereby slowing down the convergence speed. 
To address this issue, a novel initial pheromone allocation method is proposed to balance the trade-off between randomness and convergence speed in the coloy. Since the straight-line distance between two points represents the shortest path and serves as the optimal solution under ideal conditions, PFACO utilizes the Euclidean distance between the start and end points as a baseline. By computing the ratio of this baseline to the sum of the Euclidean distances from other locations to the start and end points, the proximity of these positions to the optimal solution is evaluated. This method, referred to as Adaptive Distance Pheromone Initialization (ADPI), determines the initial pheromone concentration values in the environment based on (\ref{eq:ADPI}).
\begin{equation}\label{eq:ADPI}
\tau_{ij}^{0} =  \frac{a \times Euc(ST)}{Euc(Sj) + Euc(jT)},
\end{equation}
For a more detailed explanation for $a$ in (\ref{eq:ADPI1}):
\begin{equation}\label{eq:ADPI1}
    \begin{cases}
        a=2 & \text{$d_{iT} > d_{jT} $}\\
        a=1 & \text{else} \\
    \end{cases}
\end{equation}
where $S$ denotes the starting node, $T$ denotes the terminal node, $i$ and $j$ are the available nodes, and $Euc(xy)$ denotes the Euclidean distance between $x$ and $y$. 
When $d_{iT} > d_{jT} $,indicating that node $j$ is closer to the terminal node $T$, a higher weight is assigned to node $j$ (with $a=2$). This weighting mechanism encourages the ant colony to prioritize exploration in the direction of node $j$.

This strategy provides a distribution of pheromone concentration focused on the ideal optimal solution or near-optimal solutions during the initial phase of the ant colony. Consequently, it enhances the guidance of the pheromone to the ants' exploration direction while substantially improving the algorithm's convergence speed.

\begin{algorithm}[t]
    \SetKwInOut{Input}{Input}
    \SetKwInOut{Output}{Output}
    \Input{Instance (Map ($W\times H$), start $S$ and termination $T$) and the parameters of ant colony, including the iteration $K$ and the number of ants $M$.} 
    \Output {The optimal or near-optimal solution.}
    Initialization$:$ Adaptive Distance Pheromone Initialization\;
    \For{$ w \leftarrow 1$ \KwTo $W$}  
        { 
        \For{$h \leftarrow 1$ \KwTo $H$}   
            {   
            \eIf{the node $i$ is available}{
                Calculate the Euclidean distances $Euc(ST)$, $Euc(Si)$, $Euc(iT)$, $Euc(jT)$\;
                Initialize non-uniform pheromone matrix according to (\ref{eq:ADPI})\;
            }
            {
                The pheromone concentration is $0$\;
                }
            }
        }
    Initializing the list of $Global\_Elite\_Solution$ and $Top\_Quality\_Solution$    \;
    \For{$k \leftarrow 1$ \KwTo $K$}  
        { 
        \For{$m \leftarrow 1$ \KwTo $M$}   
        {   
        \If{ant $m$ not reach the termination $T$         
        }{
            Ant $m$ selects the next node based on (\ref{eq:acoupdatepheromone}),~(\ref{eq:LTOS}) and~(\ref{eq:acotransfer}) \;
        }
            Updating the route of ant $m$ by LTOS (Section~\ref{sec:LTOS})\;
        }
        \If{all ant reach the termination $T$         
        }{
            Updating the set of $Global\_Elite\_Solution$ and $Top\_Quality\_Solution$\;
            $New\_Set=Top\_Quality\_Solution+5\times Global\_Elite\_Solution$\;
            Update pheromones matrix based on $New\_Set$ by~(\ref{eq:acoupdatepheromone}) and~(\ref{eq:LTOS})\;
            Initializing $New\_set$\;            
        }
        }
\caption{The pseudo-code of PFACO.}\label{alg:PFACO}
\end{algorithm}

\subsection{Promising Solutions Pheromone Reinforcement Strategy (PSPRS)}
To promote a significant distribution of pheromone concentration in ant colonies, the promising solutions pheromone reinforcement strategy (PSPRS) is introduced. Within the PFACO algorithm, the top $(0.1 \times M) $ global solutions identified by the ant colony are recorded in real time and designated the list of Global Elite Solutions. After each iteration, the solutions discovered by the ants are ranked according to their quality, and the top half of the solutions from the current iteration are retained as Top Quality Solutions. Each solution within the Global Elite Solutions set is then replicated five times, forming a set whose size is equivalent to half the number of ants in the population. These replicated promising solutions, together with Global Elite Solutions, contribute to updating the pheromone matrix in the environment. As shown in Line 20 to 25 of Algorithm~\ref{alg:PFACO}. 
The strategy strengthens the concentration of pheromones around high-quality solutions, promoting the long-term impact of promising solutions and accelerating pheromone accumulation on optimal solutions. Additionally, the replication mechanism not only amplifies the influence of high-quality solutions but also preserves diversity and prevents premature convergence to local optima.

By integrating PSPRS, PFACO improves global search efficiency and accelerates convergence, enhancing its ability to optimize paths in complex environments. Overall, this strategy refines pheromone distribution, strengthens the colony’s preference for high-quality solutions, and significantly improves PFACO’s performance in complex optimization tasks.

\subsection{Lookahead Turning Optimization Strategy (LTOS)}
\label{sec:LTOS}
In traditional ant colony algorithms, the state transition rule (such as~\ref{eq:acoheristic}) causes the selection process to favor adjacent nodes with shorter distances (e.g., up, down, left, and right) in addition to pheromone concentration. This preference can lead to an excessive number of turns in the global solution.
To address this shortsightedness and further refine the pheromone concentration, the proposed strategy introduces a forward-looking node reconfiguration mechanism, termed the Lookahead Turning Optimization Strategy (LTOS). This mechanism dynamically adjusts parent nodes in the local search space during the search process, reducing redundant turns and preventing excessive detours caused by local optima. 

Moreover, the pheromone updating by  
\begin{equation}\label{eq:LTOS}
\Delta \tau_{ij}^{k} = \begin{cases}
 Q/(L_{k} + Turn_{k} ), & \text{condition}\ \mathrm{b} \\
 0, & otherwise \\
\end{cases}
\end{equation}
$^{\mathrm{b}}$If ant $k$ visited edge (i,j) in its tour.

This equation considers not only the path length but also introduces the number of turning points ($Turn_{k}$) as a penalty term, effectively balancing path length and turn redundancy. By incorporating this approach, PFACO substantially reinforces the colony's capability of searching high-quality paths. Consequently, the overall path becomes both smoother and of higher quality.

\section{EXPERIMENTS}
\label{sec:experiments}

\subsection{Data sets}
\begin{figure}[t]
    \centering
    \includegraphics[width=0.6\columnwidth]{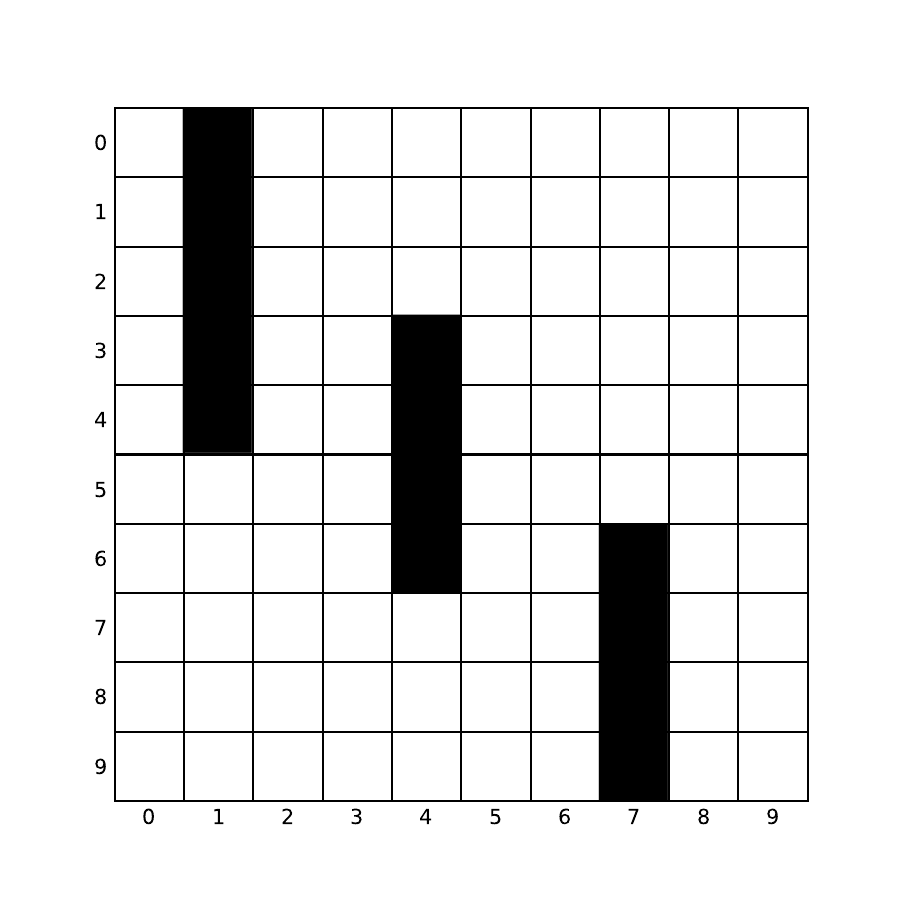}
    \caption{An example of the map dataset.}
    \label{fig:Environmentmodel}
\end{figure}

In path planning tasks, the map scale generally correlates positively with task complexity. Larger maps often involve broader spans, increasing the overall challenge.
To comprehensively evaluate the performance of the proposed PFACO algorithm across different path planning scales, this paper constructs three map datasets of different sizes based on grid-world environments referenced in ~\cite{bhardwaj2017learning} and \cite{yonetani2021path}. The datasets consist of maps sized $10\times10$, $15\times15$, and $20\times20$. For instance, the $10 \times 10$ map dataset includes one obstacle-free map, five maps with distinct obstacle configurations, and four maps with mixed obstacle. An example map is illustrated in fig.~\ref{fig:Environmentmodel}.
The space is partitioned into $N \times N $ blocks, where the black grids denote obstacles and the white grids represent traversable areas. At each time step, the agent can move in eight directions: (0,1), (0,-1), (1,1), (1,-1), (-1,1), (-1,-1), (1,0), (-1,0).

In the comparative experiments (Section~\ref{ResultsandDiscussion}), to ensure diversity in test instances, 100 instances with randomly selected start and end points are drawn from each dataset. These instances serve as the basis for evaluation, ensuring that the experimental results are both representative and comprehensive.

\subsection{Experimental setups}
All experiments in this study were conducted on a personal computer equipped with an Intel Core i7-8700 @ 3.20GHz CPU and an RTX 2080 SUPER GPU, with all algorithms implemented in Python 3.7 using PyTorch.

The selection of parameters for the comparison algorithms and PFACO algorithm discussed in this paper is primarily based on existing work \cite{liu2023learning}.
The parameters of the PFACO algorithm are set as follows: $Q=2$, $\rho=0.2$, $\alpha=1$, $\beta=3 $, and $T=2 \ or \  1$.
For AS, as described in Section~\ref{ResultsandDiscussion}, the configuration includes 30 ants and 20 iterations, denoted ``AS-30-20". 
In this study, ACO algorithm variants are labeled using the format algorithm ``name - population size - number of search iterations".
The parameters of the ACO algorithms are defined as follows: $\alpha = 1$, $\beta = 3$, the pheromone evaporation rate within $[0.1,0.4)$, and the hyperparameter $Q = 2$.

To evaluate the performance of PFACO in path planning, five ant colony algorithms were selected as baseline methods: AS, Elite AS, MMAS, NCAACO~\cite{luo2020research}, and IHMACO~\cite{liu2023improved}. Additionally, A*~\cite{hart1968formal} was included in the comparative experiments to provide a more comprehensive assessment of the overall performance of PFACO in robot path planning.

The performance of PFACO is evaluated by comparing key metrics across 100 randomly generated instances, including the average path length (\textit{AveragePath}), average time consumption per instance (\textit{Time(\%)}), number of turnings (\textit{Turning}),  success rate (\textit{SuccessRate(\%)}) and the Path Improvement Rate (\textit{PathImprove(\%)}). A shorter average path length indicates higher solution quality, while the average time consumption reflects the computational efficiency of the algorithm. 
\textit{Turning} is not an entirely independent criterion but is analyzed alongside average path length to assess algorithmic performance. Additionally, the Path Improvement Rate (\textit{PathImprove(\%)}) is used to compare the performance of the same ACO algorithm under different iteration counts and colony sizes, evaluating its convergence speed and global optimization capability. It is defined as:
\begin{equation}\label{eq:pathimprove}
 \frac{ AveragePath(a')-AveragePath(a'')}{AveragePath(a'')}
\end{equation}
Where $a'$ and $a''$ respectively represents the algorithm a with the different sets of parameter.
By analyzing how the quality of the solution changes with the increase iterations colony sizes, this study further assesses the stability of the algorithm.
Additionally, the standard deviations of the path lengths (\textit{SD-P}) and time consumption (\textit{SD-T}) are examined to evaluate the consistency and stability of the results.
To ensure the statistical significance of the findings, the Mann-Whitney U test is utilized to compare the mean differences between the algorithmic results, with the significance level set at 0.05.

To prevent excessive computational time due to potential deadlock issues when applying ACO algorithms to complex instances, a cut-off threshold is implemented for all algorithms in this section. Specifically, if a search process exceeds 120 seconds without reaching the target, it is considered unsuccessful. However, it is important to note that this mechanism may lower the success rate of ACO algorithms in solving certain instances.

\begin{table*}[ht]
\centering
\caption{The performance of PFACO and other comparative algorithms on 100 random generated instances across three different map size datasets. The evaluation metrics for these algorithms include Average Path Length (\textit{AveragePath}), Average Time Cost (\textit{Time (s)}), Average Number of Turns (\textit{Turning}), Standard Deviation of Path Length (\textit{SD-P}), Standard Deviation of Time Cost (\textit{SD-T}), Success Rate of Instance Completion (\textit{Success rate(\%)}), the degree of improvement in path planning results caused by an increased number of ants and iterations (\textit{PathImprove(\%)}), and significance results (\textit{p-value}). The best values in ACO algorithms are highlighted in bold.}
\label{table:1}
\begin{tabular}{c|cccccccc}  \toprule 
 \multicolumn{9}{c}{10 $\times$ 10} \\ \midrule
\textit{ }  & \textit{ $AveragePath$ } & \textit{Time(s)} & \textit{Turning} & \textit{SD-P} & \textit{SD-T} &\multicolumn{1}{c}{\textit{\begin{tabular}[c]{@{}c@{}}SuccessRate\\ (\%)\end{tabular}} }   & \multicolumn{1}{c}{\textit{\begin{tabular}[c]{@{}c@{}}$PathImprove$\\ (\%)\end{tabular}} } & p-value \\ \midrule

A* & 5.217    &$6.65 e^{-05}$ & 2.16  & 3.298 & $6.23e^{-05}$ & 100 & - &\textbf{0.314}  \\  \midrule
AS-15-10 & 6.157    & 0.260    & \textbf{1.18}  & 4.267   & 0.209    & 100 & -  & 0.014 \\
EliteACO-15-10        & 5.328    & 0.266    & 1.60   & 3.524   & 0.225    & 100 & -    & \textbf{0.241} \\ 
MMACO-15-10           & 5.337    & 0.277    & 1.49   & 3.528   & 0.223    & 100 & -    &\textbf{0.235} \\
NCAACO-15-10    & 5.255    & 0.986    & 1.58   & 3.404   & 0.843    & 100 & -    &\textbf{0.309} \\ 
IHMACO-15-10    & 8.809    & 2.060    & 2.03   & 6.225   & 8.863    & 94  & -    &$ 2.733e^{-07}$\\ 
PFACO-15-10          &  5.068   & \textbf{0.103}    & 1.84   & 3.231   & \textbf{0.136} & 100 &- & \textbf{0.437}\\  
AS-30-20              & 5.766    & 1.040    &  1.37  & 3.921   & 0.829    & 100 & 6.35 & \textbf{0.063} \\
EliteACO-30-20        & 5.159    & 1.040    & 1.64   & 3.277   & 0.845    & 100 & 3.17 &\textbf{0.361} \\ 
MMACO-30-20           & 5.138    & 1.080    & 1.81   & 3.284   & 0.865    & 100 & 3.73 &\textbf{0.378}\\
NCAACO-30-20    & 5.135    & 3.890    & 1.72   & 3.272   & 3.398    & 100 & 2.28 &\textbf{0.403} \\ 
IHMACO-30-20    & 8.212    & 3.460    & 2.00   & 6.013   & 10.321   & 93  & 6.77 &$4.498e^{-06}$\\  
PFACO-30-20          & \textbf{5.013} & 0.320 & 1.94 & \textbf{3.095} & 0.294   & 100 & \textbf{1.09} &- \\  
\midrule
\multicolumn{9}{c}{15 $\times$ 15} \\ \midrule
\textit{ }  & \textit{ $AveragePath $ } & \textit{Time(s)} & \textit{Turning} & \textit{SD-P} & \textit{SD-T} & \multicolumn{1}{c}{\textit{\begin{tabular}[c]{@{}c@{}}SuccessRate\\ (\%)\end{tabular}} } & \multicolumn{1}{c}{\textit{\begin{tabular}[c]{@{}c@{}}$PathImprove$\\ (\%)\end{tabular}} } & p-value\\ \midrule

A*   & 8.697    & $6.65e^{-05}$    & 2.16  & 3.298    & $2.12e^{-04}$    & 100 & - & \textbf{0.38} \\  \midrule
AS-15-10       & 12.346    & 0.937  & \textbf{2.10}  & 7.852   & 0.758    & 100 & -  & $ 2.431e^{-04}$ \\
EliteACO-15-10 & 10.410    & 0.969    & 2.48  & 6.778    & 0.789   & 100 & - & 0.046\\ 
MMACO-15-10    & 10.403    & 1.000    & 2.47  & 6.919    & 0.843    & 100 &- & 0.050\\
NCAACO-15-10   & 10.362    & 3.470    & 2.47  & 6.723   & 2.750  & 100 &- & 0.048\\ 
IHMACO-15-10    & 14.736    & 6.670   & 4.43  & 7.009   & 14.104   & 84 &- & $3.966e^{-09}$    \\  
PFACO-15-10       &  9.328   & \textbf{0.215}   & 3.27    & 5.910    & \textbf{0.173}   & 100 &- & \textbf{0.303}\\  
AS-30-20       & 11.435    & 3.840   &  2.22 & 7.246   & 3.145   & 100 & 7.37  & $3.439e^{-03}$\\
EliteACO-30-20 & 9.559    & 3.850    & 2.67  & 5.916    & 3.189    & 100  &8.17 & \textbf{0.213}\\ 
MMACO-30-20    & 9.653    & 3.940   & 2.77  & 6.137    & 3.332    & 100  &7.21 & \textbf{0.184}\\
NCAACO-30-20  & 9.681    & 3.890    & 2.70  & 6.206    & 13.032    & 100  &6.57 & \textbf{0.176}\\ 
IHMACO-30-20    & 13.301   & 7.220    & 4.11   & 7.494    & 14.269    & 82  &9.74 & $1.475e^{-05}$ \\  
PFACO-30-20       & \textbf{8.912}    & 0.695      & 3.13    & \textbf{5.412}    & 0.516   & 100  &\textbf{4.46} & - \\  
\midrule
\multicolumn{9}{c}{20 $\times$ 20} \\ \midrule
\textit{ }  & \textit{ $AveragePath $ } & \textit{Time(s)} & \textit{Turning} & \textit{SD-P} & \textit{SD-T} & \multicolumn{1}{c}{\textit{\begin{tabular}[c]{@{}c@{}}SuccessRate\\ (\%)\end{tabular}} } & \multicolumn{1}{c}{\textit{\begin{tabular}[c]{@{}c@{}}$PathImprove$\\ (\%)\end{tabular}} }  & p-value \\ \midrule

A*         & 10.982    & $2.30e^{-04}$    & 5.15  & 5.174    & $2.00e^{-04}$    & 100 &- & 0.003\\  \midrule
AS-15-10       & 19.167    & 2.21    & 3.47  & 11.523   & 1.660    & 100 &- & $9.751e^{-05}$\\
EliteACO-15-10 & 16.234    & 2.18    & \textbf{3.31}  & 9.902   & 1.638    & 100 &- & 0.025\\ 
MMACO-15-10    & 16.231    & 2.20    & 3.52  & 9.706    & 1.607    & 100 &- & 0.025\\
NCAACO-15-10     & 16.081    & 8.71     & 3.48    & 9.598    & 6.667     & 100 &- & 0.035 \\ 
IHMACO-15-10    & 20.449    & 15.20    & 5.74   & 9.605    & 26.907    & 58 &- & $2.341e^{-06}$ \\ 
PFACO-15-10 & 14.116   & \textbf{1.14}     & 4.58    & 8.560    & \textbf{0.941} & 100 &- & \textbf{0.349}\\
AS-30-20       & 18.155    & 8.59    & 3.44  & 11.038  & 6.365   & 100  &5.27 & $7.382e^{-04}$\\
EliteACO-30-20 & 15.029   & 8.66    & 3.74  & 8.962    & 6.333    & 100  &7.42 & \textbf{0.134}\\ 
MMACO-30-20    & 15.161    & 8.95    & 3.61  & 9.042    & 6.677    & 100  &6.59 & \textbf{0.114}\\
NCAACO-30-20     & 15.317    & 31.30    & 3.34    & 9.229    & 18.373    & 100  &4.75 & \textbf{0.095}\\ 
IHMACO-30-20     & 20.203    & 10.50    & 5.75   & 9.645    & 17.188    & 53  &\textbf{1.20} & $8.273e^{-06}$\\    
PFACO-30-20 & \textbf{13.664}    & 1.39     & 4.71    & \textbf{8.017}    & 1.557    & 100  &3.20 & -\\  
\bottomrule
\end{tabular}
\end{table*}

\begin{figure*}[ht]
    \centering
    \includegraphics[width=1\linewidth]{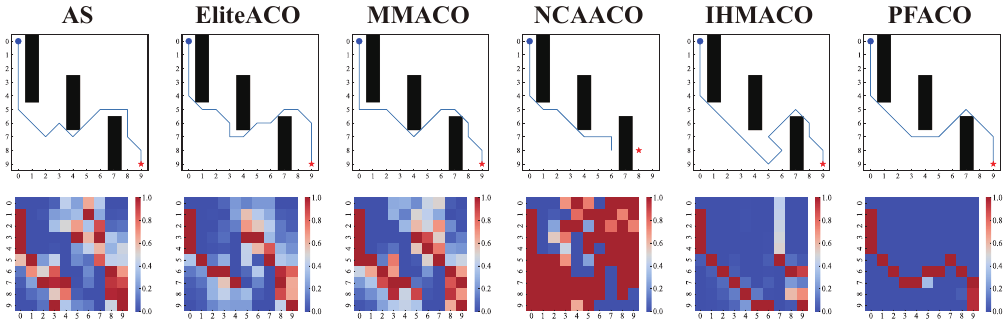}
    \caption{The path planning results and the final pheromone concentration distribution of ACO algorithms in Instance 1.}
    \label{fig:pheromonematrix}
\end{figure*}

\subsection{Results and Discussion}
\label{ResultsandDiscussion}

This section compares the performance of PFACO against AS, Elite AS, MMAS, NCAACO, and IHMACO on datasets with map sizes $10 \times 10$, $15 \times 15$, and $20 \times 20$, the results are presented in Table~\ref{table:1}. Additionally, the results of A* are included to provide a comprehensive evaluation of PFACO in path planning tasks.
The comparison is conducted across multiple performance metrics, including  \textit{AveragePath}, \textit{Time (s)}, \textit{Turning},  \textit{SD-P}, \textit{SD-T}, \textit{Success rate(\%)}, \textit{PathImprove(\%)}, and \textit{p-value}.

As shown in Table~\ref{table:1}, among all compared ACO algorithms, PFACO-30-20 achieves the smallest \textit{AveragePath} across all three map scales. Specifically, on the $10 \times 10$ map set, PFACO outperforms A* in both \textit{AveragePath} and \textit{Turning}. On the $15 \times 15$ map set, the \textit{AveragePath} of PFACO-30-20 is closest to A*, differing by only 0.215. On the $20 \times 20$ map set, PFACO-30-20 remains competitive with A* in \textit{AveragePath}, with a difference of 2.134, while its average \textit{Turning} is 0.44 lower than that of A*.
Furthermore, PFACO-30-20 exhibits the smallest \textit{SD-P} among all compared algorithms, while PFACO-15-10 ranks second, indicating that the solutions generated by PFACO are more stable than those of other algorithms. These results further confirm that PFACO consistently delivers high-quality path planning solutions across different map scales.

In terms of average computation time (\textit{Time(s)}), PFACO-15-10 achieves the shortest execution time across all experimental environments, followed by PFACO-30-20, which ranks among the top three in efficiency among the compared ant colony algorithms. Specifically, PFACO-15-10 has the smallest \textit{SD-T}, while PFACO-30-20 ranks among the top three in \textit{SD-T}, indicating that PFACO exhibits lower computational time fluctuations.

Through a comparative analysis of the improvement in \textit{AveragePath} and the magnitude of enhancement in path length before and after increasing the ant colony population size and the number of iterations, as presented in Table~\ref{table:1}, PFACO exhibits the smallest improvement while maintaining the shortest \textit{AveragePath}. This observation suggests that, in contrast to other algorithms, PFACO achieves superior path planning results with fewer ants and iterations, eliminating the necessity for large-scale populations or extensive iterations to enhance path quality significantly. These findings highlight the efficiency of PFACO in computational resource utilization, as it can obtains optimal solutions with higher efficiency, demonstrating enhanced search performance and stability in path planning tasks.


\begin{figure}[th]
    \centering
    \includegraphics[width= 1\linewidth]{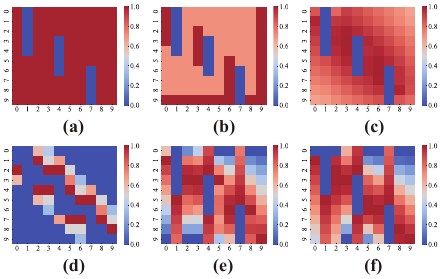}
    \caption{
    The initial pheromone concentration distribution for different strategies on Instance 1 is depicted. The vertical and horizontal axes represent the $x$ and $y$ coordinates of the grid map, respectively. The color gradient indicates pheromone concentration values, where warmer colors correspond to higher concentrations and cooler colors to lower concentrations. The concentration values range from [0, 1].
    }
    \label{fig:intial_pheromonematrix}
\end{figure}

Furthermore, Table~\ref{table:1} presents the results of the hypothesis test. Mann-Whitney U tests were performed to compare the algorithm that achieved the best \textit{AveragePath} (highlighted in bold italics) with the other algorithms. The results are shown in bold when no significant difference is detected at a significance level of 0.05.
The significance test results indicate that no statistically significant difference exists among A*, AS-30-20, EliteACO, MMACO, NCAACO, IHMACO, PFACO-15-10, and PFACO-30-20 on $10 \times 10$ map datasets. Similarly, for $15 \times 15$ map datasets, no significant difference is observed among A*, EliteACO-30-20, MMACO-30-20, NCAACO-30-20, PFACO-15-10, and PFACO-30-20. On $20 \times 20$ map datasets, EliteACO-30-20, NCAACO-30-20, PFACO-15-10, and PFACO-30-20 exhibit no statistically significant differences.
The results in Table~\ref{table:1} indicate that PFACO-15-10 and PFACO-30-20 do not exhibit significant disadvantages across all map sizes, demonstrating the strong stability of PFACO across different problem scales. 

Overall, the experimental results in Table~\ref{table:1} highlight the advantages of PFACO across multiple aspects. It achieves shorter path lengths, higher solution stability, improved computational efficiency, and reduced fluctuations. Furthermore, PFACO maintains consistent performance across varying problem scales, demonstrating robustness against search space expansion. These attributes collectively establish PFACO as a highly efficient and reliable path planning algorithm, particularly well-suited for applications requiring limited computational resources or rapid response times.

This section examines a 10-scale map (Fig.~\ref{fig:Environmentmodel}), with the start at $(0,0)$ and the goal at $(9,9)$, referred to as \textbf{Instance 1}. 
Unlike edge-confined obstacles, which increase obstacle density without significantly affecting path planning difficulty, Instance 1 features centrally located obstacles forming a `C-trap' and maze-like structure. These obstacles can mislead the algorithm’s exploration, making path planning notably more challenging. To evaluate algorithm performance, this section visualizes the final paths and pheromone distributions for an intuitive comparison. The results of AS, EliteACO, MMACO, NCAACO, IHMACO, and PFACO on Instance 1 are shown in Fig.~\ref{fig:pheromonematrix}. The top row displays the final path-planning results, while the bottom row illustrates the corresponding pheromone concentration distributions. PFACO exhibits a more concentrated pheromone distribution around the optimal or near-optimal solutions, indicating stronger convergence compared to other algorithms. In contrast, AS, EliteACO, and MMACO maintain a more dispersed pheromone distribution, even in later search stages.
The final pheromone concentration distribution of the IHMACO algorithm demonstrates the strongest convergence among the compared algorithms. However, in the latter half of the path, the presence of the L-shaped obstacle prevents it from converging near the optimal or near-optimal solution within the given iteration and ant constraints, potentially necessitating additional computational resources. As a result, its search efficiency is slightly lower than that of PFACO.
A notable case is observed in NCAACO, a mismatch between NCAACO's hyperparameters and Instance 1 may have led to ineffective transition probability calculations, causing ants to endlessly explore the environment without reaching the goal.
The superior convergence and concentrated pheromone distribution in PFACO stem from its integrated strategies: ADPI, which optimizes pheromone initialization adaptively based on specific instances; PSPRS, which accelerates global convergence; and LTOS, which enhances local path quality. These factors collectively enable PFACO to achieve faster convergence and more efficient solutions than other algorithms.

\begin{figure}[th]
    \centering
    \includegraphics[width=0.7\linewidth]{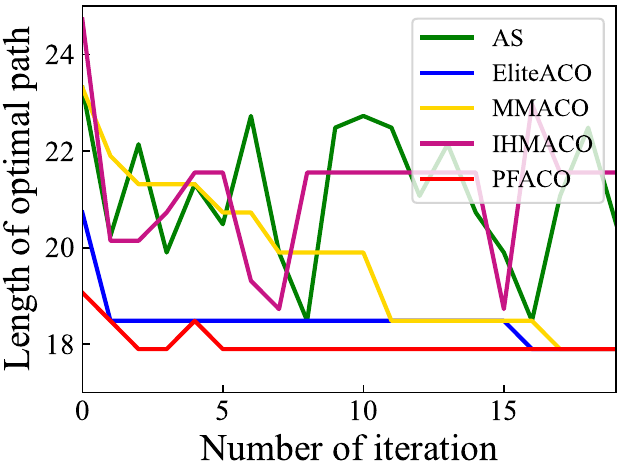}
    \caption{Path length variation curves of PFACO and other benchmark algorithms during the iteration process on Instance 1.}
    \label{fig:iteration}
\end{figure}


To demonstrate the effectiveness of ADPI, this section compares the pheromone matrices of PFACO with five other different pheromone initialization strategies.
As shown in Fig.~\ref{fig:intial_pheromonematrix}, (f) represents the pheromone initialization matrix under ADPI, compared with the following initialization methods:  (a) constant uniform distribution, (b) inverse-distance-based pheromone initialization ($\eta_{ij}$), (c) the initialization strategies used in NCAACO~\cite{luo2020research},
(d) IACO-SFLA~\cite{ pu2024improved} and (e) MsAACO~\cite{cui2024multi}. 
The comparative results indicate that before the search begins, ADPI provides the ant colony with a focused pheromone distribution, effectively filtering out the less promising search regions. This strategy enhances search precision by directing exploration toward the most valuable areas.  
In contrast, the constant uniform distribution (Fig.~\ref{fig:intial_pheromonematrix}~(a)) lacks differentiation in pheromone concentration, offering no additional guidance for subsequent searches. The inverse distance-based initialization (Fig.~\ref{fig:intial_pheromonematrix}~(b)) introduces some differentiation but provides only a vague directional bias toward the goal. The pheromone distributions in NCAACO \cite{wu2024application} (Fig.~\ref{fig:intial_pheromonematrix}~(c)) and IHMACO (Fig.~\ref{fig:intial_pheromonematrix}~(d)) are similar, with the highest pheromone concentration along the direct path between the start and goal. However, Fig.~\ref{fig:intial_pheromonematrix}~(d) IHMACO covers a broader range of valuable exploration areas than Fig.~\ref{fig:intial_pheromonematrix}~(c) NCAACO. This type of distribution is less effective in navigating complex traps, such as L-shaped or C-shaped obstacles.
The visualization results indicate that this strategy facilitates a pheromone concentration distribution centered around the desired optimal or near-optimal solution during the initial phase of PFACO. It strengthens the guidance of ant exploration, enabling them to efficiently navigate promising search regions.

\begin{table}[t]
\begin{center}
\caption{Results of 10 reproducibility experiments comparing PFACO with other comparison s algorithms on Instance 1.}
\label{table:LTOS}
\begin{tabular}{ccccccc} \toprule 
\multirow{2}{*}{Algorithm} & \multicolumn{3}{c}{Path Length} & \multicolumn{1}{c}{\multirow{2}{*}{Turning}} & \multirow{2}{*}{Time (s)} \\ \cline{2-4} 
 &Mean       & Best    & Std.     & \multicolumn{1}{c}{}                      &                           \\  
\midrule
A* &19.657  &19.657 &0.000  & 9.0 & $3.533e^{-04}$  \\ \midrule
AS & 21.499  &19.899 &1.029  & 6.1 & 4.611  \\
ElitACO & 18.802  &18.485 & 0.439 & \textbf{6.0} & 4.852  \\
MMACO & 18.724 &\textbf{17.899} & 0.444 & 6.5 & 4.909 \\
IHMACO & 21.356  &20.142 &0.690  & 8.1 & 1.462 \\
PFACO & \textbf{18.134}  & \textbf{17.899} & \textbf{0.302}  &6.8 & \textbf{1.322}\\
\bottomrule 
\end{tabular}
\end{center}
\end{table}

To illustrate the effectiveness of the PSPRS strategy, this section visualizes a comparison between PFACO and other competing algorithms by plotting the variation in solution quality achieved by the ant colonies over iterations in Instance 1.
As shown in Fig.~\ref{fig:iteration}, during the solution process for Instance 1, PFACO achieves the smallest initial solution, the fastest convergence, and the least fluctuation compared to the other benchmark algorithms. 
These results further validate the effectiveness of PSPRS, as it guides ants toward promising paths in the search process, enabling faster convergence to optimal or near-optimal solutions.
The reduced fluctuation in solution quality suggests that PSPRS stabilizes the search process, preventing erratic jumps between solutions and promoting more consistent exploration of the solution space. 
This indicates that PSPRS effectively enhances the ant colony's preference for high-quality solutions, thereby improving its performance in solving complex optimization problems.

To evaluate the effectiveness of LTOS, this section presents the results of 10 reproducibility experiments conducted with PFACO and other algorithms on Instance 1. 
As shown in Table~\ref{table:LTOS}, although PFACO does not achieve the lowest turning count among all algorithms, it consistently yields the shortest average path length and the best path length across repeated experiments, with the smallest standard deviation.
These results indicate that despite not minimizing turns, PFACO excels in both path quality and stability. Specifically, LTOS effectively reduces unnecessary turns, leading to smoother and more reliable solutions, as reflected by the minimal variation in path length (as shown in $Std.$ of $Path Length$). This demonstrates that PFACO successfully balances turning and path length, achieving more stable and efficient path planning compared to other algorithms.

\section{CONCLUSION}
\label{conclusion}

This paper proposes the Pheromone Focusing Ant Colony Optimization (PFACO) algorithm, which optimizes the concentration of pheromones to focus on regions around potential optimal or near-optimal solutions, thereby improving solution quality and convergence speed. The performance of PFACO is evaluated against several benchmark algorithms across different map scales. Experimental results demonstrate that PFACO consistently outperforms other comparison algorithms, particularly on smaller maps, where its solution quality surpasses that of the deterministic path planning algorithm.
Furthermore, this paper validates the effectiveness of ADPI, PSPRS, and LTOS. 
These findings establish PFACO as a valuable addition to the family of ACO-based methods. However, like other ant colony-based algorithms, PFACO incurs higher computational costs on large-scale maps. Future work will focus on improving its scalability and efficiency in dynamic and complex environments to mitigate these limitations, and on extending the proposed method to a broader range of application domains \cite{guo2025modeling,wei2025multiobjective}.



\section*{ACKNOWLEDGMENT}

This study was supported in part by the Shanghai Municipal Science and Technology Major Project (No. 2021SHZDZX0103).

\bibliography{references} 
\bibliographystyle{IEEEtran}

\end{document}